\documentclass{article}
\usepackage{iclr2027_conference,times}
\usepackage{placeins}
\usepackage[T1]{fontenc}
\usepackage[dvipsnames]{xcolor}
\usepackage{amsmath,amssymb,graphicx}
\usepackage{booktabs,multirow,gensymb,xspace,textcomp,pifont}
\usepackage{caption,subcaption}
\usepackage{abbreviations}
\usepackage[accsupp]{axessibility}
\usepackage[hyphens]{url}
\usepackage[pagebackref]{hyperref}
\usepackage[capitalize]{cleveref}
\crefname{section}{Sec.}{Secs.}
\Crefname{section}{Section}{Sections}
\crefname{table}{Tab.}{Tabs.}
\Crefname{table}{Table}{Tables}
\let\cite\citep
\definecolor{Coral}{rgb}{1, 0.47, 0.24}

\newcommand{\OURS}{QuadTok\xspace}

\newcommand{\cmark}{\ding{51}}
\newcommand{\xmark}{\ding{55}}
\title{\OURS: Quadtree Visual Tokenizer for Autoregressive Image Generation}
\author{%
  \textbf{Yucheng Mao\textsuperscript{*}, Zeyuan Chen\textsuperscript{*}, Xiaojun Shan, Xiang Zhang}\\
  \textbf{Divyansh Srivastava, Bingnan Li, Zhuowen Tu}\\[0.6ex]
  \normalfont University of California, San Diego\\[0.3ex]
  \normalfont\small\textsuperscript{*}Equal Contribution.
}

\iclrfinalcopy
\begin{document}

\maketitle
\lhead{Preprint.} % Override the conference-publication status for arXiv.

\noindent\begin{minipage}{\textwidth}
\vspace{-2mm}
\centering
\includegraphics[width=\linewidth]{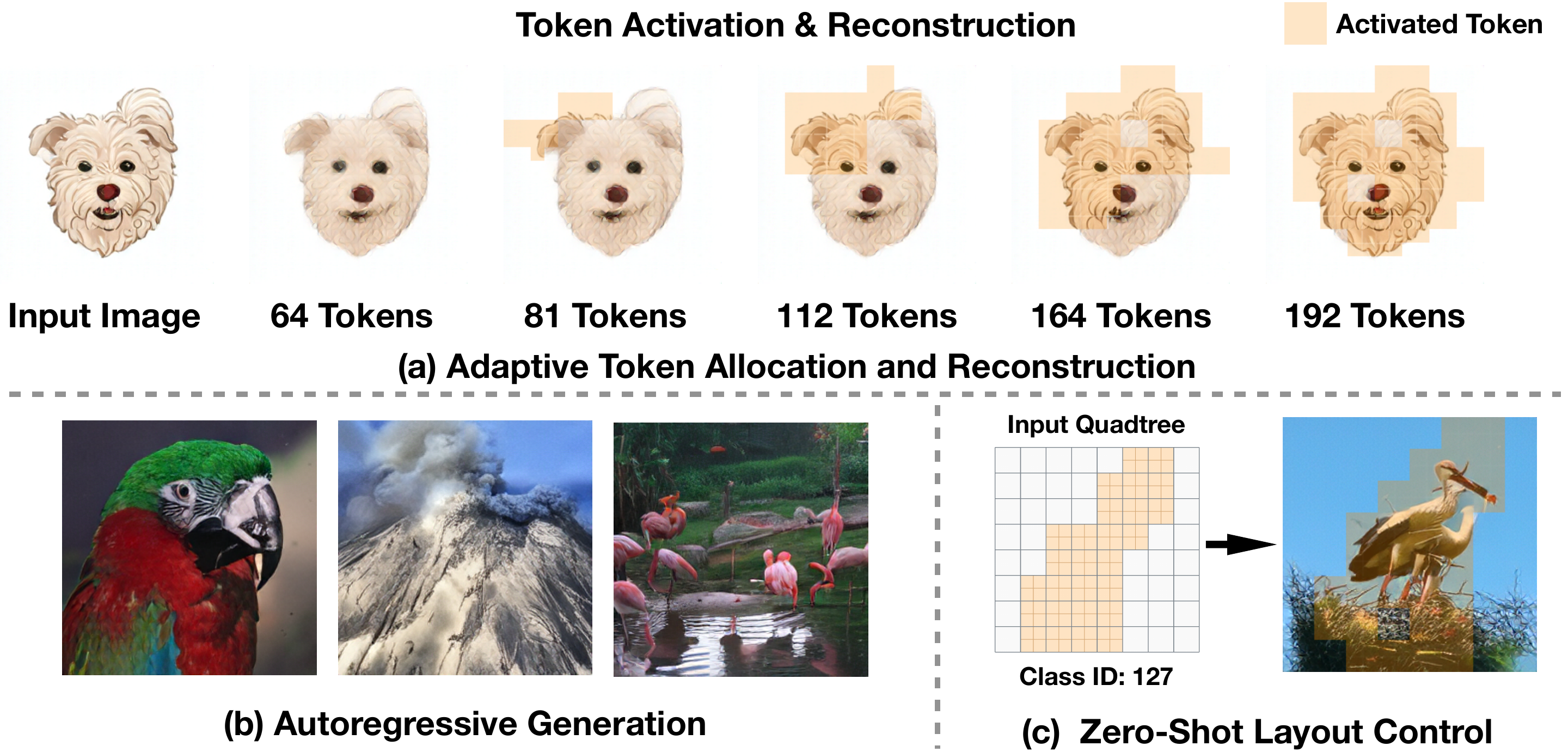}
\captionsetup{hypcap=false}
\captionof{figure}{\textbf{Adaptive token allocation with \OURS}: (a) Progressive reconstruction using the \OURS tokenizer: starting from a coarse reconstruction with 64 tokens, additional tokens progressively improve reconstruction quality. (b) Autoregressive image generation with the \OURS generator: tokens are generated following the quadtree topology, enabling coarse-to-fine synthesis. (c) Zero-shot spatial layout controlled generation: a user or LLM-specified spatial layout is translated into a quadtree, and the \OURS generator predicts visual codes conditioned on this topology.}
\label{fig:teaser}
\vspace{-2mm}
\end{minipage}\par\medskip

\begin{abstract}
  We introduce \OURS, a novel framework for visual tokenization and autoregressive image generation. Compared to traditional approaches using 2D grids or 1D token sequences, we propose a hierarchical quadtree structure, bridging the gap between 2D spatial binding and 1D sequence-level flexibility. The \OURS tokenizer dynamically allocates representational capacity to visually intricate areas while leaving homogeneous regions at a coarse resolution. Compared with a fixed 256-token grid, our ImageNet-trained tokenizer saves approximately \textbf{10\%} of tokens on ImageNet and \textbf{9\%} when transferred zero-shot to COCO dataset, while maintaining comparable reconstruction fidelity. Furthermore, the natural causality introduced by the tree structure seamlessly enables autoregressive image generation. Conditioned on a quadtree topology supplied before generation, our 947M GPT-style generative model achieves a \textbf{2.08} gFID on the ImageNet $256 \times 256$ benchmark. Additionally, leveraging the strong spatial correlation preserved by the quadtree structure, the \OURS generator enables zero-shot spatially controlled image generation capabilities. Code: \url{https://github.com/myc634/QuadTok}.
  
  % \keywords{Autoregressive Visual Generation \and Visual Tokenization \and Hierarchical Image Representation}
\end{abstract}

% \begin{abstract}
%   We introduce \OURS, a quadtree based visual tokenizer that dynamically allocates tokens across the image plane to account for spatially varying complexities. Traditional visual tokenizers rely on patch-based tokens or rigid fixed-length representations, which limit their ability to handle the non-uniform information distribution in natural images. In contrast, \OURS leverages a hierarchical tree structure to distribute tokens across regions, enabling multi-resolution and fine-grained control over token allocation. It improves tokenization efficiency while achieving strong reconstruction fidelity. Furthermore, the hierarchical structure naturally aligns with the causal formulation of autoregressive generative models, supporting coarse-to-fine generation that starts with global structure and progressively refines local details. Experimental results on ImageNet and COCO across resolutions demonstrate that \OURS achieves state-of-the-art image generation quality.
%   \keywords{Autoregressive Visual Generation \and Visual Tokenization \and Hierarchical Image Representation}
% \end{abstract}
\section{Introduction} \label{sec:intro}

Natural images exhibit highly non-uniform spatial information density: regions containing intricate textures and high-frequency patterns carry significantly more information than smooth or homogeneous areas. Ideally, a visual tokenizer should allocate representational capacity according to this spatial variability. However, classic visual tokenizers~\cite{oord2018neuraldiscreterepresentationlearning,sun2024autoregressive} compress images into rigid 2D grids with a fixed token budget, distributing tokens uniformly across the image regardless of local complexity. As a result, complex regions are often under-represented while simple regions are over-allocated with redundant tokens.

To overcome the rigidity of grid-based tokenization, recent work has explored 1D visual tokenizers~\cite{shen2025cat,yu2024imageworth32tokens,duggal2024adaptive,miwa2025onedpieceimagetokenizermeets,wang2026visualgranularitygeneration,bachmann2025flextokresamplingimages1d,chen2026soft}. These methods model images as sequential token streams and allow for variable length, offering greater flexibility in adapting the token budget to image complexity. Nevertheless, this flexibility comes at the cost of spatial awareness. Because token allocation occurs along the sequence dimension rather than the image plane, these approaches remain largely spatially agnostic. Consequently, they often lack explicit correspondence between tokens and image patches, which can degrade pixel-level fidelity and make them less suitable for downstream tasks that require spatially grounded representations and precise localization.

In this work, we propose \textit{\OURS}, a novel quadtree-based visual tokenizer that enables adaptive token allocation while preserving explicit spatial correspondence. Instead of relying on a rigid grid or a purely sequential representation, \OURS represents images using a hierarchical quadtree structure that dynamically subdivides spatial regions based on local complexity. This design bridges the gap between existing paradigms: it retains the spatial grounding of traditional 2D tokenizers while introducing the sequence-level flexibility of 1D token representations. However, one key challenge remains: \textit{How can we select which regions to refine based on their reconstruction benefit?}

To tackle this, we propose \textit{Region-Wise Complexity Guidance}, a data-driven strategy to select a content-adaptive tree for each image based on reconstruction benefit. Furthermore, we identify that maintaining the hierarchical causality and explicit spatial relationships between tree nodes is crucial during the tokenization process. To enforce this structural integrity, we introduce the \textit{Kinship Causal Mask}. This mechanism dictates a strict coarse-to-fine information flow along the quadtree hierarchy, allowing the global image structure to be established before localized refinement occurs within spatially bounded regions.

Empowered by our \textit{Region-Wise Complexity Guidance}, the \OURS tokenizer achieves an rFID of 1.46 and a PSNR of 20.37 on ImageNet with 230 tokens on average, saving approximately 10\% relative to a fixed 256-token grid. Transferred zero-shot to COCO, the same tokenizer uses 232 tokens on average, saving approximately 9\%. Furthermore, leveraging the causality inherent in the quadtree structure, the \OURS 947M autoregressive generator achieves a gFID of 2.08, compared with 2.15 for the 1.4B RandAR-XXL model~\cite{pang2025randar}. We also show a consistent scaling behavior on our generator across different model scales. Furthermore, guided by the explicit quadtree topology, \OURS generator enables novel zero-shot capabilities such as spatially controllable generation.

Our contributions are summarized as follows:
\begin{itemize}
    \item We propose \textit{\OURS}, a quadtree-based visual tokenizer that replaces fixed-grid representations with spatially adaptive token allocation. Our approach preserves explicit spatial correspondence while enabling flexible token budgets driven by local visual complexity.

    \item We introduce two key mechanisms to support adaptive tokenization: (i) \textit{Region-Wise Complexity Guidance} strategy selecting regional refinement from reconstruction benefit; (ii) a \textit{Kinship Causal Mask} enforcing hierarchical dependencies among tree nodes during tokenization.

    \item Leveraging the hierarchical causality of quadtrees, \OURS achieves competitive image generation quality while enabling zero-shot spatially controllable image synthesis conditioned on quadtree layouts supplied before generation.
    
\end{itemize}

% To ensure that hierarchical tokens maintain strict spatial relationships when processed by Transformers, we introduce the \textit{Kinship Causal Mask}. This mechanism enforces a coarse-to-fine information flow along the quadtree hierarchy, allowing global image structure to be established before localized refinement occurs within spatially bounded regions. Furthermore, we propose \textit{Region-Wise Complexity Guidance}, a data-driven strategy to determine the optimal tree topology for each image. Guided by reconstruction-based complexity estimation, this mechanism selectively activates high-resolution tokens only in visually complex regions. As a result, \OURS achieves high reconstruction fidelity while substantially reducing the token budget.
\section{Related Work}
\label{sec:related_work}

\subsection{Visual Tokenization}
Visual tokenizers compress images into representations for generative modeling. VQ-VAE~\cite{oord2018neuraldiscreterepresentationlearning} learns discrete codebooks, and VQ-GAN~\cite{esser2021taming} adds adversarial training to improve perceptual reconstruction. Later work improves codebook design and tokenizer architectures~\cite{lee2022autoregressive,mentzer2023finite,yu2023language,zhao2024image}. Beyond fixed spatial grids, TiTok~\cite{yu2024imageworth32tokens} represents images as compact 1D sequences. FlexTok~\cite{bachmann2025flextok} and One-D-Piece~\cite{miwa2025onedpieceimagetokenizermeets} support variable sequence lengths, while GigaTok~\cite{xiong2025gigatok} uses semantic regularization to improve reconstruction and downstream generation as tokenizers scale. \OURS focuses on assigning discrete tokens to regions within an image while retaining an explicit spatial correspondence.

\paragraph{Content-adaptive tokenization.}
CAT~\cite{shen2025cat} adapts representation capacity to image complexity by using caption-derived scores to select an image-wide compression ratio in a nested continuous VAE. Its variable-length latents support image synthesis with a diffusion transformer. \OURS instead allocates discrete tokens at different spatial resolutions within each image, preserving explicit correspondence between tokens and regions. During reconstruction, image content determines which regions to refine; during generation, an autoregressive model predicts visual codes under a topology supplied before sampling. We compare with CAT's published ImageNet $512\times512$ generation results in \cref{tab:generation_512}, alongside other high-resolution generators.

\subsection{Autoregressive Image Generation}
Autoregressive image models differ in both their latent representation and prediction unit~\cite{lee2022autoregressive, li2024autoregressive, shan2026cyclegen}. LlamaGen~\cite{sun2024autoregressive} predicts discrete grid tokens sequentially, RandAR~\cite{pang2025randar} changes their generation order, and VAR~\cite{tian2024visual} predicts successive resolution scales. DPAR~\cite{srivastava2025dpardynamicpatchificationefficient} uses an entropy model to group discrete tokens into variable-length patches, reducing the sequence processed by its global transformer. We compare with DPAR as a related approach to spatial token aggregation. \OURS represents regions directly as quadtree nodes and generates their discrete codes in breadth-first order conditioned on the given tree.

Continuous-token methods provide another approach to autoregressive synthesis. xAR~\cite{ren2025beyond} generalizes the prediction unit to entities such as cells and uses flow matching with noisy contexts. FlowAR~\cite{ren2025flowar} combines scale-wise autoregression with flow matching. \OURS models variable-length discrete sequences with explicit regional correspondence, allowing generation to be conditioned on a supplied spatial topology.

\subsection{Quadtree Representations in Vision}
\label{sec:related_quadtree}
Quadtrees recursively partition two-dimensional space and have been used for image processing, compression, and adaptive visual computation~\cite{Wu_2018,tang2022quadtreeattentionvisiontransformers,ronen2023visiontransformersmixedresolutiontokenization,zhao2025qliptextalignedvisualtokenization}. \OURS incorporates this hierarchy into a generative visual tokenizer. Its topology-aware attention preserves structured interactions among tokens, and its discrete representation supports image reconstruction and autoregressive synthesis under a supplied spatial topology.

\section{Method}
\label{sec:method}

\begin{figure*}[t!]
\centering
\includegraphics[width=0.98\linewidth]{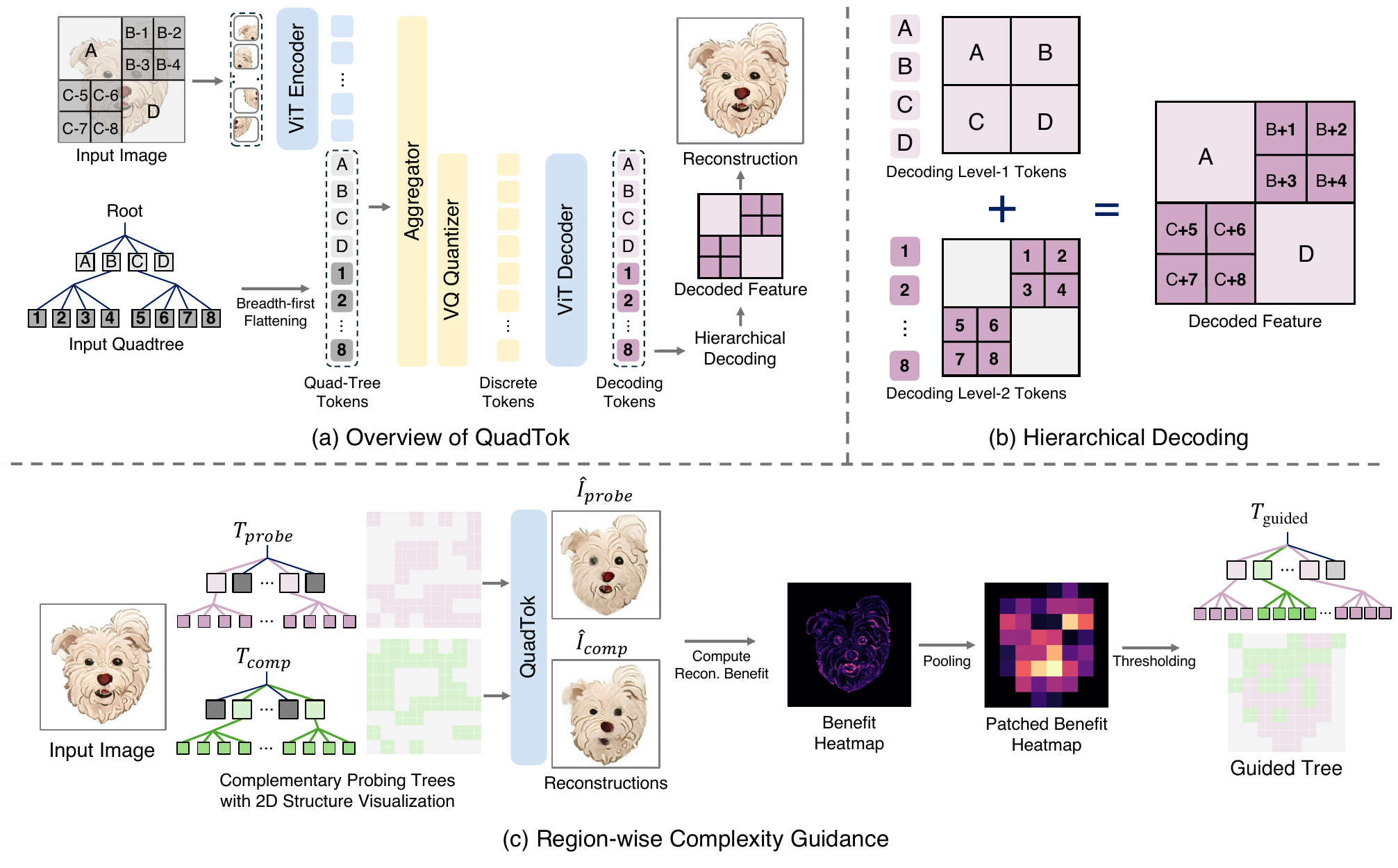}
\vspace{-2mm}
\caption{\textbf{(a) Overview of QuadTok.} An image is serialized into a 1D token sequence from a quadtree via breadth-first traversal. ViT features are aggregated into quadtree tokens, quantized, and decoded hierarchically. \textbf{(b) Hierarchical decoding.} Coarse tokens recover global structure; fine tokens refine local regions. \textbf{(c) Region-wise Complexity Guidance.} Complementary probing trees estimate per-region reconstruction benefit to build a content-adaptive quadtree.}
\vspace{-1mm}
\end{figure*}

\subsection{\OURS Tokenizer}

We introduce \OURS, a visual tokenizer that represents images using an adaptive quadtree structure. Unlike conventional methods that rely on fixed grids, \OURS dynamically allocates tokens across regions while maintaining explicit correspondence between tokens and the image patches.

\subsubsection{Quadtree Representation}

\OURS represents an image using a hierarchical quadtree decomposition, where each node corresponds to a spatial region and is treated as a visual token. A node is defined by the attribute tuple $(l, p)$, where $l$ denotes the tree level and $p$ the spatial index. Level $l=1$ represents coarse regions of the image, while level $l=2$ corresponds to higher-resolution subdivisions. Each parent node may expand into four child nodes covering its spatial quadrants. Our default configuration uses two levels; we also evaluate a three-level extension in Secs.~4.1 and~4.2.

To interface this hierarchical structure with Transformer architecture, the quadtree is serialized into a 1D sequence using breadth-first search (BFS) traversal. This ordering first lists all coarse tokens and then their corresponding higher-resolution child tokens, naturally aligning with coarse-to-fine generation.

\subsubsection{Image Encoding}

Given an input image $I$, we first extract dense visual features using a Vision Transformer(ViT)~\cite{dosovitskiy2021imageworth16x16words} encoder $F = \mathrm{Encoder}(I)$ where $F \in \mathbb{R}^{N \times D}$ denotes grid-based patch features.

Each quadtree node is represented by a \emph{learnable instruction token} that encodes its spatial location and tree level. For node $i$ with tree attributes $(l_i, p_i)$, the instruction embedding is $q_i = \mathrm{Emb}_{l_i}(p_i)$

The sequence of instruction tokens $Q = [q_1,\cdots,q_M]$ is concatenated with the dense image features and processed by a Transformer module named \emph{Aggregator}: $[F^{\prime} \parallel Z] = \mathrm{Aggregator}([F \parallel Q])$

Through self-attention, each region-indexed instruction token reads from the dense, globally contextualized ViT feature map. The resulting embeddings $Z=[z_1,\cdots,z_M]$ form continuous representations of quadtree nodes associated with their respective spatial regions.

\subsubsection{Vector Quantization}
To obtain discrete tokens compatible with autoregressive modeling, the node embeddings are quantized using a learned codebook $\mathcal{C} = \{e_k\}_{k=1}^K$: $\mathrm{Quant}(z_i) = \arg\min_{e_k \in \mathcal{C}} \| z_i - e_k \|_2$

The resulting discrete token sequence constitutes the image representation.

\subsubsection{Hierarchical Decoding}
To reconstruct the image from the discrete token sequence, we reverse the tokenization process through a coarse-to-fine hierarchical decoding procedure.

The discrete tokens are first mapped back to continuous embeddings using the codebook and combined with their corresponding structural instruction embeddings. For token $i$ associated with node $(l_i, p_i)$, we have ${z}_i=e_{y_i} + q_i$ where $e_{y_i}$ denotes the embedding of the discrete token index and $q_i$ is the node instruction embedding encoding the tree level and spatial position. A Transformer-based \emph{Decoder} then processes the embedding sequence as $\tilde{Z}=[\tilde{z}_1,\cdots,\tilde{z}_M]$.

Unlike standard grid-based decoders that directly reshape the output sequence into a dense spatial grid, \OURS reconstructs the image according to the quadtree hierarchy.

Each processed node feature $\tilde{z}_i$ is first projected into a spatial patch using a level-specific projection module: $C_{l_i}(p_i) = U_{l_i}(\tilde{z}_i)$ where $U_l(\cdot)$ denotes the unpatching operator for level $l$, producing a spatial feature patch corresponding to region $p$. The patches from the same level are placed at their corresponding spatial locations on a canvas to form a residual feature map $C_l$. We then aggregate the feature map level-by-level as $H_{l+1} = F^{\mathrm{up}}_{l+1} (H_l) + C_{l+1}$ where $H_0 = 0$, and $F^{\mathrm{up}}_{l+1}$ is a learned upsampling module for level $l+1$.

This process starts with a coarse global representation using level-1 tokens and then progressively refines local details using higher resolution tokens. Finally, the highest-resolution feature map ($H_2$ in two-level configuration) is projected to RGB space, yielding the final reconstruction $\hat{I} \in \mathbb{R}^{H\times W \times 3}$.

\subsubsection{Tree-Structure Sampling}

Although the tokenizer can operate on arbitrary quadtree structures, training with a fixed fully expanded tree would cause the model to rely on a deterministic token layout. To improve robustness to variable tree structures, we employ a probabilistic tree sampling strategy during training.

Instead of always using a fully expanded quadtree, we randomly sample valid subtrees for each training iteration. All level-$1$ nodes are always retained to preserve the global image layout, while each parent node independently expands into its four child nodes with probability $\gamma$. When a parent node is not expanded, its spatial region remains represented solely by the coarse-level token.

This stochastic sampling exposes the tokenizer to diverse tree configurations and sequence lengths during training. As a result, the model learns to produce consistent representations under both coarse and refined token layouts, enabling it to robustly process the adaptive tree structures used during complexity-guided tokenization (to be introduced in \cref{sec:guide}).

\subsection{Kinship Causal Attention}
\label{subsec:ae}

Flattening the quadtree into a 1D sequence removes the explicit hierarchical constraints of the original structure. Without additional restrictions, standard self-attention would allow tokens to interact globally, potentially breaking the spatial correspondence between tokens and their associated image regions. To preserve the hierarchical topology, we introduce \emph{Kinship Causal Attention}, a topology-aware masking strategy that constrains attention according to the quadtree structure.

During encoding, the \emph{Aggregator} Transformer processes the concatenated sequence $[F \parallel Q]$ of ViT feature tokens $F$ and quadtree instruction tokens $Q$. Attention is asymmetric: ViT tokens attend bidirectionally among themselves, and quadtree tokens attend to all ViT tokens, but not vice versa. Tree tokens thus draw on globally contextualized image features, while their region and level embeddings specify the regions they represent.

During decoding, the \emph{Decoder} Transformer processes the quantized visual tokens together with their corresponding quadtree instruction tokens. In both the encoding and decoding stages, the self-attention interactions \emph{among quadtree tokens} are governed by our proposed \textit{Kinship Causal Mask}, which enforces a strict coarse-to-fine dependency consistent with the quadtree hierarchy.

This mask is used within the tokenizer, whereas the autoregressive generator uses a \textbf{standard causal mask} over the BFS token sequence. For the two-level tokenizer, the Kinship Causal Mask follows three structural rules:

\begin{itemize}
    \item \textbf{Parent $\leftrightarrow$ Parent.} 
    Coarse-level tokens attend to each other bidirectionally to establish the global image layout.

    \item \textbf{Child $\rightarrow$ Parent.} 
    Fine-level tokens attend to all parent tokens to obtain global structural context, while parent tokens are prevented from attending to child tokens.

    \item \textbf{Child $\leftrightarrow$ Sibling.} 
    Fine-level tokens attend only to other child tokens that share the same parent node, restricting refinement within each spatial region.
\end{itemize}

\begin{figure}[t]
\vspace{-2mm}
    \centering
    \begin{minipage}[c]{0.30\linewidth}
        \centering
        \includegraphics[width=\linewidth]{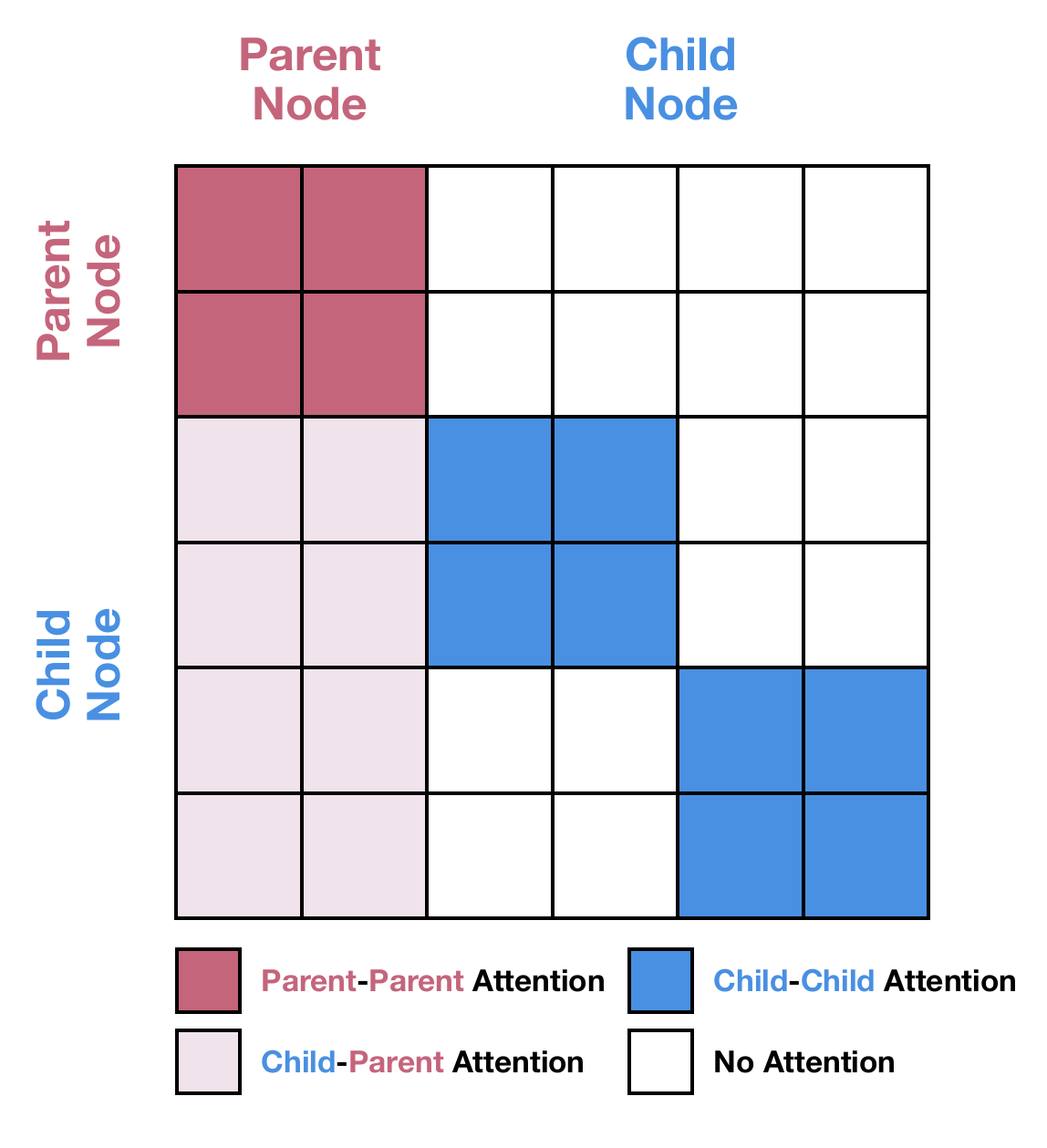}
    \end{minipage}\hfill
    \begin{minipage}[c]{0.67\linewidth}
        \centering
        \includegraphics[width=\linewidth]{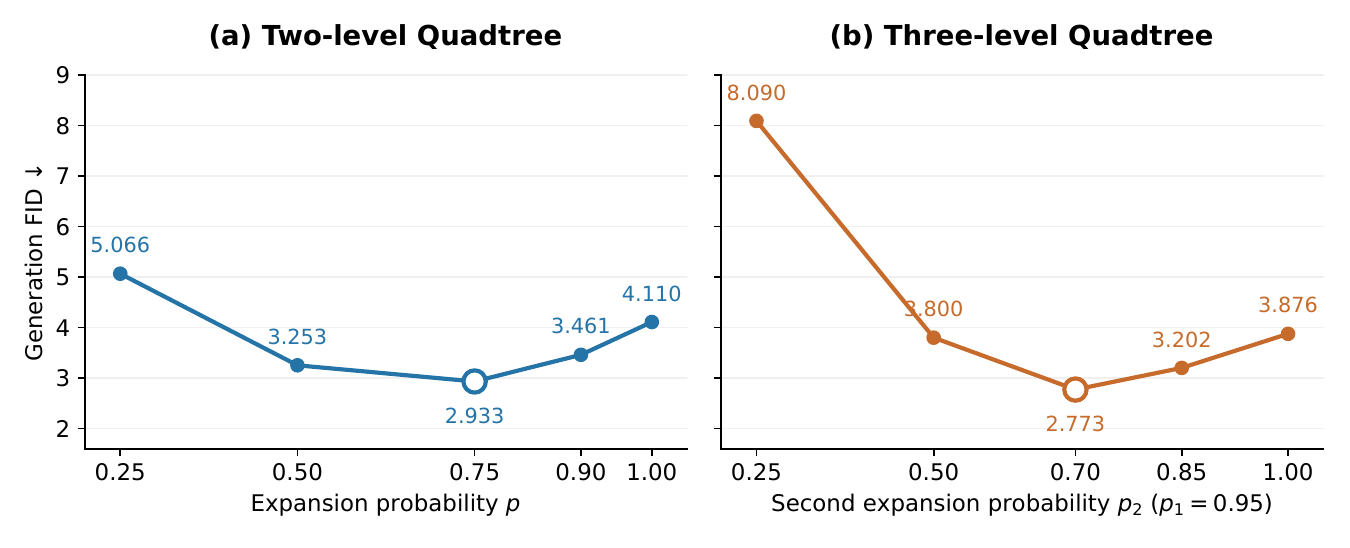}
    \end{minipage}
    \par\vspace{-0.3em}
    \captionsetup{font=small,justification=raggedright,singlelinecheck=false,skip=4pt}
    \begin{minipage}[t]{0.38\linewidth}
        \hyphenpenalty=10000
        \captionof{figure}{\textbf{Kinship causal mask.} Child, parent, and sibling attention preserves hierarchical spatial correspondence.}
        \label{fig:sib_mask}
    \end{minipage}\hfill
    \begin{minipage}[t]{0.62\linewidth}
        \hyphenpenalty=10000
        \captionof{figure}{\textbf{Generation quality versus Quadtree expansion.} Two-level $p$ (left) and three-level $p_2$ with $p_1=0.95$ (right). Open markers indicate the lowest evaluated FID. }
        \label{fig:expansion_fid}
    \end{minipage}
\vspace{-2mm}
\end{figure}

\cref{fig:sib_mask} illustrates the two-level pattern. The same rule extends to deeper trees: a node can attend to all nodes at coarser levels and to siblings sharing its immediate parent at the same level, but not to finer-level nodes. The coarsest level retains full bidirectional attention. Thus, level-3 nodes attend to all level-1 and level-2 nodes and bidirectionally to their level-3 siblings.

These constraints impose a coarse-to-fine flow and restrict same-level interactions among tree tokens to siblings. The ablation in \cref{sec:ablation_kinshipmask} evaluates their effect on reconstruction and generation.

\subsection{Region-wise Complexity Guided Tokenization}
\label{sec:guide}
While the tokenizer can process arbitrary tree structures, we must determine which regions should be refined for a given image. Intuitively, regions containing complex textures require finer-level tokens, whereas homogeneous regions can remain coarse. We measure regional complexity using \emph{reconstruction benefit}. A region is considered complex if reconstructing it with finer-level tokens significantly improves perceptual fidelity. To estimate this efficiently, we perform a probing procedure. We randomly expand half of the coarse regions to form a probing tree and construct its complementary tree where the remaining regions are expanded. This balanced design allows each region to appear once in an expanded state and once in a collapsed state, enabling efficient estimation of the regional reconstruction benefit for all regions simultaneously. The image is reconstructed under both tree structures, producing two outputs. We compute a spatial LPIPS~\cite{zhang2018unreasonableeffectivenessdeepfeatures} loss map between each reconstruction and the original image $I$. For a region $r$, let $E_r^{\mathrm{collapsed}}$ and $E_r^{\mathrm{expanded}}$ denote the loss maps from the probing reconstructions in which that region is collapsed and expanded, respectively. Its estimated regional reconstruction benefit is $b_r = \operatorname{AvgPool}_{r}\!\left(E_r^{\mathrm{collapsed}}-E_r^{\mathrm{expanded}}\right)$. Here, $\operatorname{AvgPool}_{r}$ averages the loss difference over the spatial extent of region $r$. A positive $b_r$ indicates that expansion reduces reconstruction error relative to the input image.

Regions with $b_r>\tau$ are expanded into four child nodes, where $\tau$ controls the refinement threshold. We then tokenize the image using the selected content-adaptive tree.

This strategy allocates tokens to visually complex regions while keeping simple regions at coarse resolution, enabling efficient adaptive tokenization.

\subsection{Quadtree-Based Autoregressive Image Generation}
Given a quadtree structure $T$ supplied before generation, we synthesize its visual token sequence $Y=[y_1,\cdots,y_M]$ autoregressively: $p(Y | c, T) = \prod_{i=1}^{M} p(y_i | y_{<i}, c, t_{\leq i})$ where $c$ denotes the conditioning signal (\eg a class label) and $t_i=(l_i,p_i)$ represents the tree attributes of node $i$. The topology is selected before sampling, but node instructions are presented in BFS order. At step $i$, the model can use only the current and preceding node instructions, not future nodes.

We parameterize this distribution using a LLaMA~\cite{llama1}-style decoder-only Transformer with a standard causal attention mask. Each visual-token prediction depends only on preceding visual codes and the supplied conditioning information; the tokenizer's bidirectional parent and sibling interactions are not used in the generator. Structural instruction embeddings encode the quadtree geometry, which specifies both the tree level and spatial position of each token, enabling the model to condition generation on spatial hierarchy.

% During inference, generation proceeds sequentially following the BFS ordering of the tree. Because coarse tokens appear earlier in the sequence, the model first establishes the global structure of the image and then progressively refines local details.

During inference, tokens are generated sequentially in BFS order. Coarse tokens appear first, establishing the global structure, after which finer tokens progressively refine local details and textures.

The generator predicts visual codes under the supplied topology, enabling layout-controlled synthesis from user-defined or algorithmically constructed trees.

\section{Experiments}
\label{sec:exps}

\subsection{\OURS Image Reconstruction}

\begin{table}[t]
\vspace{-1mm}
\caption{\textbf{Image reconstruction on ImageNet-1K and COCO at $256\times256$.} \#Tokens reports dataset-specific mean sequence lengths for \OURS. $f$ denotes the LlamaGen downsampling factor. High-resolution results are provided in \cref{sec:high_resolution_512}. -- denotes unavailable values.}
\label{tables/img_recon_imgnet_coco}
\vspace{-2.5mm}
\centering
\setlength{\tabcolsep}{0.45em}
\resizebox{0.95\linewidth}{!}{%
\begin{tabular}{clcrrrrrr}
\toprule
\multirow{2}{*}{\textbf{Type}} & \multirow{2}{*}{\textbf{Tokenizer}} & \multirow{2}{*}{\textbf{Codebook}} & \multicolumn{3}{c}{\textbf{ImageNet-1K}} & \multicolumn{3}{c}{\textbf{COCO}} \\
\cmidrule(lr){4-6}\cmidrule(lr){7-9}
 & & & \textbf{\#Tokens} & \textbf{rFID}$\downarrow$ & \textbf{PSNR}$\uparrow$ & \textbf{\#Tokens} & \textbf{rFID}$\downarrow$ & \textbf{PSNR}$\uparrow$ \\
\midrule
\multirow{4}{*}{2D} & Taming VQ-GAN~\cite{esser2021taming} & 16384 & 256 & 4.98 & 19.40 & 256 & 19.29 & 19.57 \\
 & MaskGIT VQ-GAN~\cite{chang2022maskgit} & 1024 & 256 & 2.28 & $-$ & -- & $-$ & $-$ \\
 & LlamaGen ($f=16$)~\cite{sun2024autoregressive} & 16384 & 256 & 2.19 & 20.79 & 256 & 8.11 & 20.42 \\
 & LlamaGen ($f=8$)~\cite{sun2024autoregressive} & 16384 & 1024 & 0.59 & 24.45 & 1024 & 4.19 & 24.20 \\
\midrule
\multirow{7}{*}{1D} & TiTok-S~\cite{yu2024imageworth32tokens} & 4096 & 128 & 1.71 & 17.80 & 128 & 9.22 & 17.27 \\
 & FlexTok~\cite{bachmann2025flextok} & 64000 & 256 & 1.08 & 17.70 & -- & $-$ & $-$ \\
 & One-D-Piece-L~\cite{miwa2025one} & 4096 & 256 & 1.08 & 19.04 & 256 & 8.04 & 17.94 \\
 & GigaTok-S-S~\cite{xiong2025gigatok} & 16384 & 256 & 1.01 & 20.74 & -- & $-$ & $-$ \\
 & GigaTok-S-B~\cite{xiong2025gigatok} & 16384 & 256 & 0.89 & 20.93 & -- & $-$ & $-$ \\
 & GigaTok-B-L~\cite{xiong2025gigatok} & 16384 & 256 & 0.81 & 21.21 & -- & $-$ & $-$ \\
 & GigaTok-XL-XXL~\cite{xiong2025gigatok} & 16384 & 256 & 0.79 & 21.65 & -- & $-$ & $-$ \\
\midrule
\multirow{2}{*}{Quadtree} & \OURS & 16384 & 230 & 1.46 & 20.37 & 232 & 7.88 & 19.30 \\
 & \hspace{0.6em}-- 3-level & 16384 & 989 & 0.770 & 23.31 & 1012 & 3.92 & 23.83 \\

\bottomrule
\end{tabular}%
}
\vspace{-0.5mm}
\end{table}

\paragraph{Implementation Details.}
We train the \OURS tokenizer on the ImageNet-1K~\cite{russakovsky2015imagenetlargescalevisual} training set at $256\times256$ resolution, following LlamaGen's data augmentation pipeline~\cite{sun2024autoregressive}. The encoder, aggregator, and decoder use Transformer backbones with the structural constraints described in Sec.~3. Architecture and training details are in Appendix~\ref{sec:tokenizer_training}.

\paragraph{Evaluation Protocol.}
We evaluate on the ImageNet-1K and MS-COCO~\cite{lin2015microsoftcococommonobjects} validation sets, applying the ImageNet-trained tokenizer to COCO without fine-tuning. We report PSNR, rFID~\cite{heusel2018ganstrainedtimescaleupdate}, and mean token count.%  \cref{tables/img_recon_imgnet_coco} compares fixed-grid LlamaGen (downsampling factors 16 and 8) with representative 1D and 2D tokenizers.

\paragraph{Results.}
 \cref{tables/img_recon_imgnet_coco} shows the results. Compared with LlamaGen at downsampling factor 16, content-adaptive allocation achieves lower rFID with fewer tokens, at the cost of lower PSNR. This tradeoff persists under zero-shot transfer to COCO (\cref{tables/img_recon_imgnet_coco}).

\paragraph{Reconstruction with a deeper quadtree.}
To evaluate the extension beyond two levels, we add a $32\times32$ token grid to the default $8\times8 \rightarrow 16\times16$ hierarchy at the same image resolution. The three-level tokenizer retains the shared codebook and Kinship Mask rules, with each expanded node contributing four children. The reconstruction improvements in \cref{tables/img_recon_imgnet_coco} demonstrate that the representation supports finer spatial subdivision on both datasets. Depth and token budget increase together in this experiment. High-resolution ($512{\times}512$) reconstruction is reported in Appendix~\ref{sec:high_resolution_512}.

\subsection{\OURS Generation}
\label{sec:gen_exp}

\paragraph{Implementation Details.}
\OURS uses a decoder-only autoregressive generator following LlamaGen~\cite{sun2024autoregressive}. The two-level generator family comprises 111M (B), 343M (L), 697M (XL), and 947M (XXL) models, and the three-level generator contains 344M parameters. Generation predicts visual codes conditioned on a topology specified before sampling. For the unconditional topology setting, we sample random trees independently of the image content to be generated. Generator training and sampling details are provided in Appendix~\ref{sec:generator_training}.

\begin{table}[t]
\vspace{-2mm}
\caption{\textbf{Class-conditional generation on ImageNet evaluated at $256\times256$.} These are system-level comparisons with different training and sampling settings. Size denotes generator parameter count. $^{\ddagger}$: native-256 LlamaGen results. xAR and FlowAR use continuous 2D latents.}
\label{tab:main_results}
\centering
\setlength{\tabcolsep}{0.45em}
\resizebox{0.9\linewidth}{!}{%
\begin{tabular}{llccccc}
\toprule
\textbf{Tokens} & \textbf{Model} & \textbf{Size} & \textbf{gFID}$\downarrow$ & \textbf{IS}$\uparrow$ & \textbf{Prec.}$\uparrow$ & \textbf{Rec.}$\uparrow$ \\
\midrule
1D & TiTok-S-128--MaskGIT~\cite{yu2024imageworth32tokens} & 287M & 1.97 & 281.80 & -- & -- \\
 & TiTok-S-128--AR$^{*}$~\cite{yu2024imageworth32tokens} & 775M & 4.90 & 191.71 & 0.77 & 0.56 \\
 & One-D-Piece-L$^{*}$~\cite{miwa2025onedpieceimagetokenizermeets} & 775M & 2.99 & 235.10 & 0.81 & 0.59 \\
 & FlexTok d18-d28~\cite{bachmann2025flextok} & 1.33B & 1.86 & -- & -- & -- \\
 & GigaTok-B-L~\cite{xiong2025gigatok} & 1.4B & 2.03 & 238.52 & 0.80 & 0.63 \\
 & GigaTok-XL-XXL~\cite{xiong2025gigatok} & 1.4B & 1.98 & 256.76 & 0.81 & 0.62 \\
\midrule
2D & VAR-d20~\cite{tian2024visual} & 600M & 2.57 & 302.60 & 0.83 & 0.56 \\
 & VAR-d30~\cite{tian2024visual} & 2.0B & 1.92 & 323.10 & 0.82 & 0.59 \\
 & xAR-L~\cite{ren2025beyond} & 608M & 1.28 & 292.50 & 0.82 & 0.62 \\
 & xAR-H~\cite{ren2025beyond} & 1.1B & 1.24 & 301.60 & 0.83 & 0.64 \\
 & FlowAR-L~\cite{ren2025flowar} & 589M & 1.90 & 281.40 & 0.83 & 0.57 \\
 & FlowAR-H~\cite{ren2025flowar} & 1.9B & 1.65 & 296.50 & 0.83 & 0.60 \\
 & RandAR-L~\cite{pang2025randar} & 343M & 2.55 & 288.82 & 0.81 & 0.58 \\
 & RandAR-XL~\cite{pang2025randar} & 775M & 2.22 & 314.21 & 0.80 & 0.60 \\
 & RandAR-XXL~\cite{pang2025randar} & 1.4B & 2.15 & 321.97 & 0.79 & 0.62 \\
 & LlamaGen-B$^{\ddagger}$~\cite{srivastava2025dpardynamicpatchificationefficient} & 111M & 5.46 & 193.61 & 0.84 & 0.46 \\
 & LlamaGen-L$^{\ddagger}$~\cite{srivastava2025dpardynamicpatchificationefficient} & 343M & 3.80 & 248.30 & 0.83 & 0.52 \\
 & LlamaGen-XL$^{\ddagger}$~\cite{srivastava2025dpardynamicpatchificationefficient} & 775M & 3.39 & 227.10 & 0.81 & 0.54 \\
 & DPAR-B~\cite{srivastava2025dpardynamicpatchificationefficient} & 120M & 3.98 & 250.62 & 0.83 & 0.49 \\
 & DPAR-L~\cite{srivastava2025dpardynamicpatchificationefficient} & 352M & 2.93 & 269.34 & 0.81 & 0.56 \\
 & DPAR-XL~\cite{srivastava2025dpardynamicpatchificationefficient} & 789M & 2.67 & 281.65 & 0.82 & 0.56 \\
\midrule
Quadtree & \OURS-B & 111M & 4.41 & 205.61 & 0.85 & 0.46 \\
 & \OURS-L & 343M & 2.93 & 226.69 & 0.82 & 0.54 \\
 & \hspace{0.6em}-- 3-level-L & 344M & 2.77 & 231.70 & 0.82 & 0.55 \\
 & \OURS-XL & 697M & 2.23 & 262.38 & 0.807 & 0.60 \\
 & \OURS-XXL & 947M & 2.08 & 273.35 & 0.82 & 0.59 \\
\bottomrule
\end{tabular}%
}
% \vspace{-1mm}
\end{table}

% Continuous AR comparison table temporarily omitted.

\subsubsection{ImageNet Generation Benchmark.}
We evaluate class-conditional ImageNet generation using gFID~\cite{heusel2018ganstrainedtimescaleupdate}, Inception Score (IS)~\cite{salimans2016improvedtechniquestraininggans}, and precision/recall~\cite{kynkaanniemi2019improved}. In \cref{tab:main_results}, \OURS improves gFID over native-256 LlamaGen at comparable scales and stays competitive with larger discrete-token generators, with continuous-token xAR and FlowAR as additional references. These are cross-setting comparisons; our design choices are isolated in the ablations (\cref{sec:ablation}). At reduced token budgets, \OURS also achieves lower generation latency than LlamaGen configured for $256\times256$ images at matched model scales; the timing protocol is provided in Appendix~\ref{sec:efficiency_details}. We further validate \OURS at native $512{\times}512$; these high-resolution reconstruction and generation results are provided in Appendix~\ref{sec:high_resolution_512}.

% \OURS-L (789M parameters) achieves an FID of $2.01$, outperforming all comparable 1D tokenizer baselines: One-D-Piece-L (FID $2.99$) and TiTok-S-128 with vanilla AR decoding (FID $4.90$), while remaining competitive with FlexTok (FID $1.86$, $1.5$B parameters). Against 2D tokenizer baselines at similar scale, \OURS-L surpasses LlamaGen-XL (FID $2.62$, 775M), RandAR-XL (FID $2.22$, 775M), and VAR-600M (FID $2.57$). At the mid-range scale, \OURS-B (352M) achieves FID $2.23$, outperforming LlamaGen-L at a comparable 343M parameter count (FID $3.07$) and matching RandAR-XL which requires over twice the parameters. These results confirm that \OURS's quadtree-structured token sequence provides a strong inductive bias for autoregressive generation, enabling better sample quality per parameter.
% \subsubsection{Scaling Behavior.}
% For two-level generation, gFID decreases from 4.419 at 111M to 2.933 at 343M and 2.081 at 947M. IS increases from 205.61 to 273.3 across these operating points. The three-level 344M model achieves gFID 2.773 and IS 231.70, demonstrating generation with a deeper topology. Its comparison with the two-level 343M model is not a matched-budget depth ablation; the random-tree expansion curves will characterize their respective quality--budget tradeoffs.

\paragraph{Generation with a deeper quadtree.}
We evaluate generation under supplied three-level topologies using random trees formed by recursively expanding nodes into four children. The results in \cref{tab:main_results}, together with Sec.~4.1, validate the representation for both reconstruction and generation beyond two levels. \cref{fig:expansion_fid} examines the two hierarchies at different expansion probabilities, using the same CFG schedule and holding other sampling settings fixed within each sweep. Both curves show that increasing expansion initially improves gFID, but further expansion reverses the trend. The best evaluated quality therefore occurs before maximal expansion.

\subsection{Zero-Shot Spatial Layout Control with \OURS}
\label{sec:zero_shot_layout}

\begin{figure}[t]
\vspace{-1mm}
    \centering
    \includegraphics[width=1.0\linewidth]{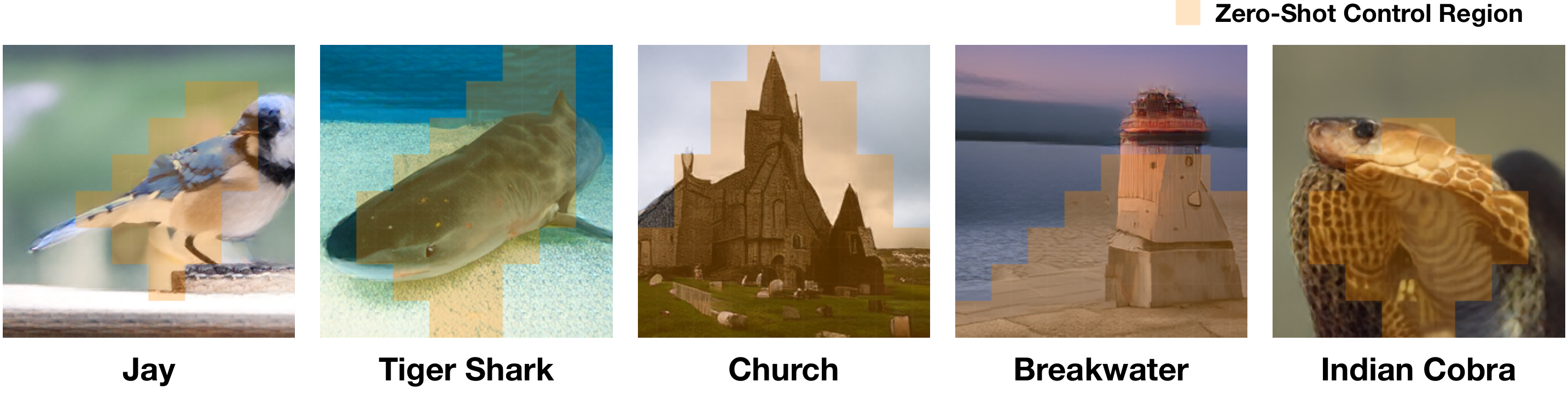}
    \caption{\textbf{Zero-shot spatial layout control.} Highlighted regions specify where the input quadtree is refined. Generated subjects align with these regions using the frozen ImageNet-trained model.}
    \label{fig:layout_plan}
\vspace{-3mm}
\end{figure}

\paragraph{Layout interface and demonstrations.}
A user or LLM~\cite{yang2025qwen3,bai2025qwen3} specifies a subject region, which is converted into a quadtree by expanding cells inside it and keeping coarse nodes elsewhere. The class label sets the category, and the topology is fixed before generation. We reuse the ImageNet-trained generator and tokenizer, both frozen, without layout-specific training. \cref{fig:teaser}(c) and \cref{fig:layout_plan} show how this steers subjects toward the expanded regions.

\paragraph{Quantitative evaluation.}
We evaluate four equally sized target half-planes (left, right, top, bottom) to test both horizontal and vertical steering. Evaluation stays within the ImageNet classes, so it tests zero-shot control of the existing generator. Each prescribed topology is compared with random topologies at the same token budget, which isolates the effect of spatial allocation. These experiments use predefined layouts. Details are in Appendix~\ref{sec:layout_eval_details}.

\paragraph{Localization metrics.}
We measure target-region hit rates using the intended-class Grad-CAM center of mass, the Grounding DINO box center, and the centroid of a box-prompted SAM~2 mask. A hit requires the estimated subject location to fall in the requested half-plane. All generated images contribute to the denominator, with missed detections counted as failures for box and mask metrics.

\begin{table}[t]
\caption{\textbf{Zero-shot spatial control with a frozen ImageNet-trained \OURS{} generator.} Prescribed vs.\ random topologies at 192 tokens, 1,000 images each. Grad-CAM, Box, and Mask: \% of subjects in the target half-plane ($\uparrow$). FID-50cls ($\downarrow$): against 2,500 class-matched real images.}
\vspace{-2mm}
\centering

\label{tab:layout_control}
\vspace{-1mm}
\small
\setlength{\tabcolsep}{0.6em}
\resizebox{0.85\linewidth}{!}{%
\begin{tabular}{llrrrr}
\toprule
Target & Topology & Grad-CAM & Box & Mask & FID-50cls \\
\midrule
\multirow{2}{*}{Left} & Random & 29.6 & 47.5 & 51.1 & 35.03 \\
 & Prescribed & \textbf{58.3} \textcolor{ForestGreen}{(+28.7)} & \textbf{51.8} \textcolor{ForestGreen}{(+4.3)} & \textbf{59.0} \textcolor{ForestGreen}{(+7.9)} & 35.15 \textcolor{BrickRed}{(+0.12)} \\
\addlinespace
\multirow{2}{*}{Right} & Random & 70.4 & 51.0 & 48.5 & 35.03 \\
 & Prescribed & \textbf{87.9} \textcolor{ForestGreen}{(+17.5)} & \textbf{58.1} \textcolor{ForestGreen}{(+7.1)} & \textbf{55.7} \textcolor{ForestGreen}{(+7.2)} & 34.88 \textcolor{ForestGreen}{(-0.15)} \\
\addlinespace
\multirow{2}{*}{Top} & Random & 33.9 & 40.7 & 42.9 & 35.03 \\
 & Prescribed & \textbf{55.4} \textcolor{ForestGreen}{(+21.5)} & \textbf{47.0} \textcolor{ForestGreen}{(+6.3)} & \textbf{47.8} \textcolor{ForestGreen}{(+4.9)} & 34.97 \textcolor{ForestGreen}{(-0.06)} \\
\addlinespace
\multirow{2}{*}{Bottom} & Random & 66.1 & 58.9 & 56.7 & 35.03 \\
 & Prescribed & \textbf{83.6} \textcolor{ForestGreen}{(+17.5)} & \textbf{64.6} \textcolor{ForestGreen}{(+5.7)} & \textbf{62.5} \textcolor{ForestGreen}{(+5.8)} & 35.05 \textcolor{BrickRed}{(+0.02)} \\
\bottomrule
\end{tabular}
}
\vspace{-1mm}
\end{table}

\paragraph{Results.}
Prescribed layouts improve all three localization hit rates across the four directions (\cref{tab:layout_control}), demonstrating that spatial token allocation can steer subject placement under a fixed token budget. The corresponding FID-50cls changes range from $-0.15$ to $+0.12$ points relative to the matched random-topology baseline.

\subsection{Ablation Studies}
\label{sec:ablation}

\subsubsection{Effectiveness of the Kinship Causal Mask}
\label{sec:ablation_kinshipmask}

We ablate the \textit{Kinship Causal Mask} (\cref{subsec:ae}) by training a baseline \OURS with unconstrained bidirectional attention over all flattened tree tokens, and study its effect on spatial correlation, reconstruction, and generation (\cref{tables/ablation_sibling}).

\paragraph{Spatial correlation.}
To probe token--region binding, we perturb a single fine-level token ($l{=}2$) with a random codebook index and compare the perturbed and unperturbed reconstructions. With the mask, the resulting difference remains localized to the corresponding spatial patch; without it, the difference spreads across the image. This spatial correspondence supports the zero-shot spatial control in \cref{sec:zero_shot_layout}.

\paragraph{Reconstruction and generation.}
This spatial isolation is also reflected quantitatively (\cref{tables/ablation_sibling}). Without the mask, the tokenizer degenerates toward a sequential 1D code: rFID is slightly lower ($1.46{\rightarrow}1.41$), but PSNR drops by over a point ($20.37{\rightarrow}19.16$), trading spatial fidelity for perceptual realism. The gap widens downstream: with the same 697M generator and CFG, removing the mask raises gFID from $2.23$ to $3.17$, showing that spatially structured tokens matter for AR generation.

% Kinship-mask perturbation visualization omitted from the paper layout.
% \begin{figure}[t]
% \vspace{-6mm}
%     \centering
%     \includegraphics[width=1.0\linewidth]{figs/sibling_mask_ablation.pdf}
%     \vspace{-6mm}
%     \caption{\textbf{Ablation of the \textit{Kinship Causal Mask} via token perturbation.} For each example, a single fine-level token (highlighted) is replaced with a random codebook index and the pixel-wise difference map is computed. Without the mask (right), unconstrained attention allows perturbation errors to propagate globally. With the mask (left), the discrepancy remains localized within the corresponding spatial patch, demonstrating strong spatial correspondence between tokens and image regions.}
%     \label{fig:sib_mask_ablation}
% \vspace{-2mm}
% \end{figure}

\begin{table}[t]
% \vspace{-2mm}
    \begin{minipage}[t]{0.46\linewidth}
        \captionof{table}{\textbf{Ablation of the \textit{Kinship Causal Mask}.} }
\vspace{-1.5mm}
\setlength{\tabcolsep}{5pt}
\label{tables/ablation_sibling}
\centering
\resizebox{\linewidth}{!}{%
\begin{tabular}{@{} cccc @{}}
    \toprule
    \textbf{Kinship Mask}        & \textbf{rFID} $\downarrow$ & \textbf{PSNR} $\uparrow$ & \textbf{gFID} $\downarrow$ \\
    \midrule
    \xmark & 1.41          & 19.16         & 3.17                       \\
    \cmark   & 1.46          & 20.37         & 2.23                       \\
    \bottomrule
\end{tabular}%
}

    \end{minipage}
    \hfill
    \begin{minipage}[t]{0.46\linewidth}
        \captionof{table}{\textbf{Effect of \textit{region-wise complexity guidance} on ImageNet reconstruction.}}
\vspace{-1.5mm}
\setlength{\tabcolsep}{11pt}
\label{tables/img_recon_guide}
\centering
\resizebox{\linewidth}{!}{%
\begin{tabular}{@{} lccc @{}}
\toprule
\textbf{Quadtree} & \textbf{\#Tokens} $\downarrow$ & \textbf{rFID} $\downarrow$ & \textbf{PSNR} $\uparrow$ \\ 
\midrule
Random & 253  & 1.601 & 19.40 \\
Full & 320  & 1.50 & 20.39 \\
Guided & \textbf{230} & \textbf{1.46} & \textbf{20.37}  \\ 
\bottomrule
\end{tabular}
}
    \end{minipage}
\vspace{-1mm}
\end{table}

\subsubsection{Region-wise Complexity Guidance Strategy.} 

\paragraph{Does guidance help?}
We compare quadtree constructions on ImageNet-1K reconstruction (\cref{tables/img_recon_guide}). A uniformly expanded (\emph{Full}) tree reconstructs well but is token-heavy ($320$), and a \emph{Random} tree of similar size is worse (rFID $1.60$ at $253$ tokens). Guided by \textit{Region-wise Complexity Guidance} (\cref{sec:guide}), \OURS concentrates tokens on visually complex regions and matches \emph{Full} fidelity (rFID $1.46$, PSNR $20.37$) with $28\%$ fewer tokens ($230$). We adopt LPIPS-based guidance with $\tau{=}0.05$ as the default; the full probing-metric and threshold ablation is provided in Appendix~\ref{sec:guidance_ablation}.

\section{Conclusion}
\label{sec:conclusion}
We present \OURS{}, a quadtree-based tokenizer that enables spatially adaptive, variable-length visual representations. By organizing tokens in a quadtree structure, \OURS{} allocates tokens according to local complexity while preserving spatial correspondence between tokens and image regions which bridges adaptive tokenization and structured spatial representations. Experiments show that \OURS{} achieves strong reconstruction and generation performance while enabling zero-shot spatial layout control. While this work focuses on class-conditional generation, \OURS{} is compatible with multimodal pipelines and text-to-image generation, which we leave for future work.

\subsection*{Acknowledgment}

\noindent \textbf{Acknowledgement.} This work is supported by U.S. National Science Foundation Award IIS-2433768 and IIS-2127544. This work also used DeltaAI at National Center for Supercomputing Applications (NCSA) through allocation CIS262057 from the Advanced Cyberinfrastructure Coordination Ecosystem: Services \& Support (ACCESS) program, which is supported by U.S. National Science Foundation grants \#2138259, \#2138286, \#2138307, \#2137603, and \#2138296.
We thank Haiyang Xu and Ethan Armand for helpful discussions.

{
    \small
    \bibliographystyle{iclr2027_conference}
    \bibliography{main}
}

\clearpage
\appendix
\raggedbottom
\section{Additional Qualitative Examples}
\label{sec:supp_qualitative}

\cref{fig:supp_gen_birds,fig:supp_gen_canids,fig:supp_gen_panda_valley,fig:supp_gen_volcano_geyser} show class-conditional generation at $256\times256$ resolution. Each grid contains 25 samples from one ImageNet class; we include the complete grids without selecting individual images.

% Token-allocation visualization (former Figure 7) omitted from the paper layout.
% \begin{figure}[htbp]
% \vspace{-2mm}
%     \centering
%     \includegraphics[width=\linewidth]{figs/token_activate_map.pdf}
%     \caption{\textbf{Samples of adaptive token allocation.} As image complexity increases (left to right), \OURS activates more quadtree tokens (highlighted with orange overlay), resulting in higher token usage and finer spatial subdivision in detailed regions.}
%     \label{fig:token_act}
% \vspace{-3mm}
% \end{figure}

\noindent\begin{minipage}{\textwidth}
    \centering
    \begin{minipage}[t]{0.49\textwidth}
        \centering
        \includegraphics[width=\linewidth]{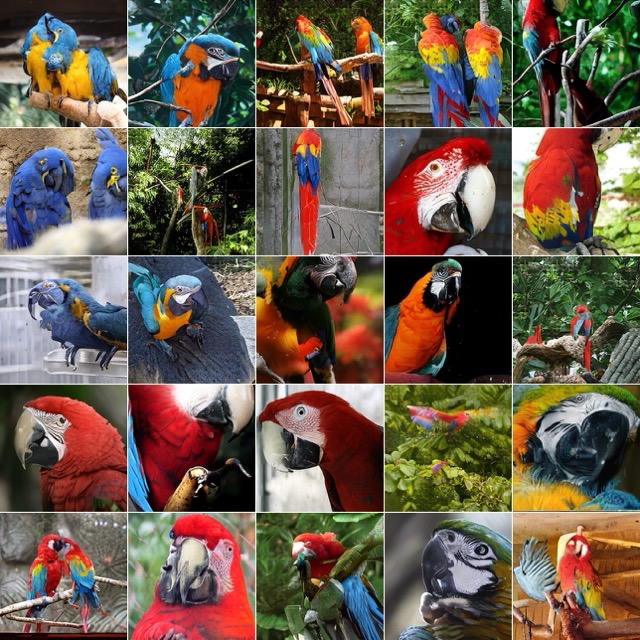}
    \end{minipage}\hfill
    \begin{minipage}[t]{0.49\textwidth}
        \centering
        \includegraphics[width=\linewidth]{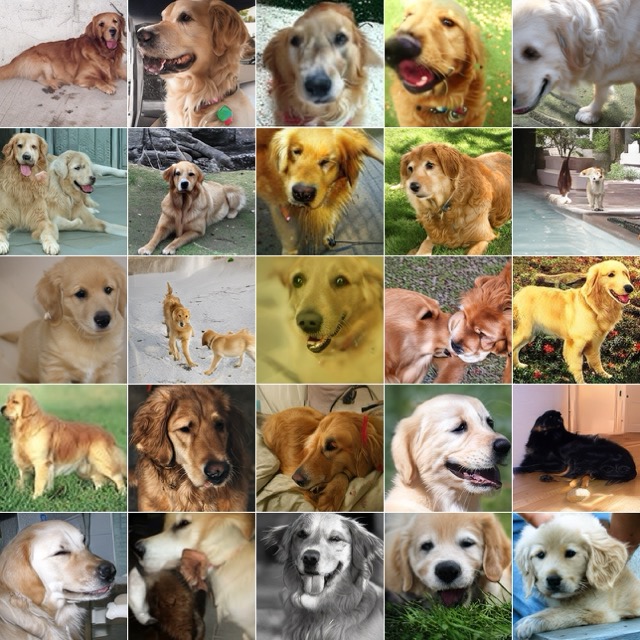}
    \end{minipage}
    \captionsetup{hypcap=false}
    \captionof{figure}{Uncurated class-conditional generation on ImageNet-256. Left: macaw (class 88). Right: golden retriever (class 207). Each grid contains 25 samples.}
    \label{fig:supp_gen_birds}
\end{minipage}\par\medskip

\noindent\begin{minipage}{\textwidth}
    \centering
    \begin{minipage}[t]{0.49\textwidth}
        \centering
        \includegraphics[width=\linewidth]{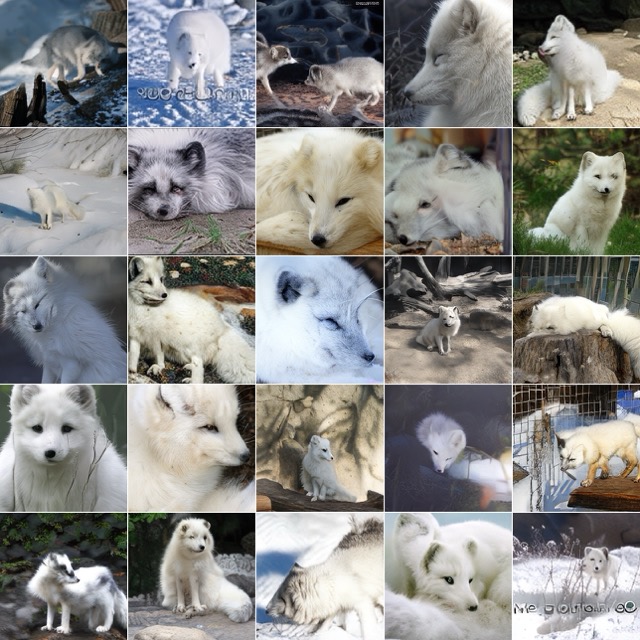}
    \end{minipage}\hfill
    \begin{minipage}[t]{0.49\textwidth}
        \centering
        \includegraphics[width=\linewidth]{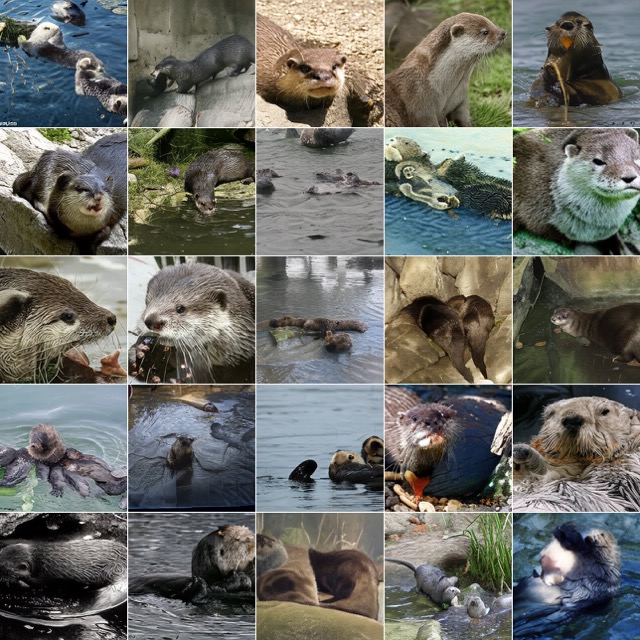}
    \end{minipage}
    \captionsetup{hypcap=false}
    \captionof{figure}{Uncurated class-conditional generation on ImageNet-256. Left: Arctic fox (class 279). Right: otter (class 360). Each grid contains 25 samples.}
    \label{fig:supp_gen_canids}
\end{minipage}\par\medskip

\noindent\begin{minipage}{\textwidth}
    \centering
    \begin{minipage}[t]{0.49\textwidth}
        \centering
        \includegraphics[width=\linewidth]{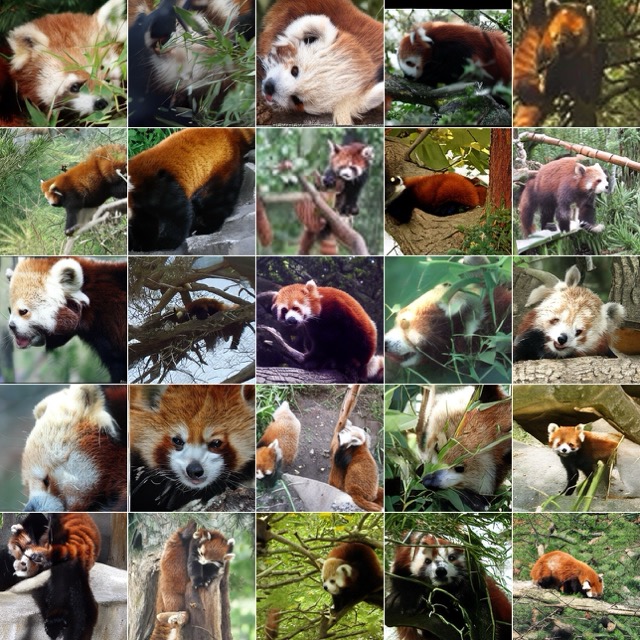}
    \end{minipage}\hfill
    \begin{minipage}[t]{0.49\textwidth}
        \centering
        \includegraphics[width=\linewidth]{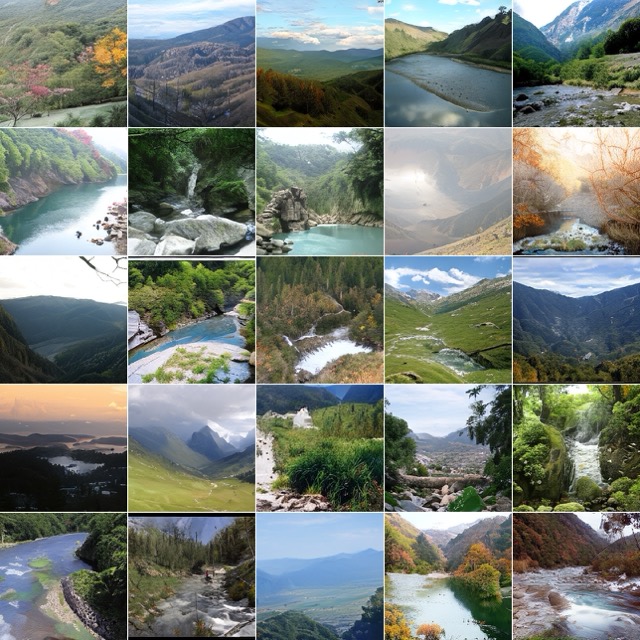}
    \end{minipage}
    \captionsetup{hypcap=false}
    \captionof{figure}{Uncurated class-conditional generation on ImageNet-256. Left: lesser panda (class 387). Right: valley (class 979). Each grid contains 25 samples.}
    \label{fig:supp_gen_panda_valley}
\end{minipage}\par\medskip

\noindent\begin{minipage}{\textwidth}
    \centering
    \begin{minipage}[t]{0.49\textwidth}
        \centering
        \includegraphics[width=\linewidth]{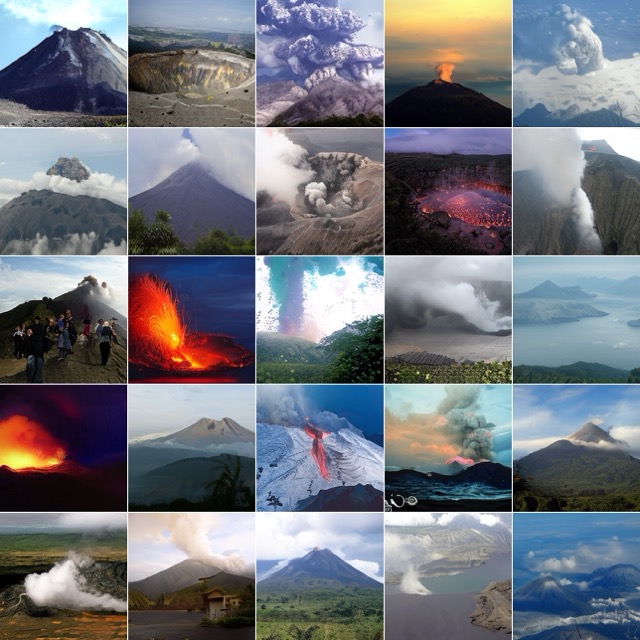}
    \end{minipage}\hfill
    \begin{minipage}[t]{0.49\textwidth}
        \centering
        \includegraphics[width=\linewidth]{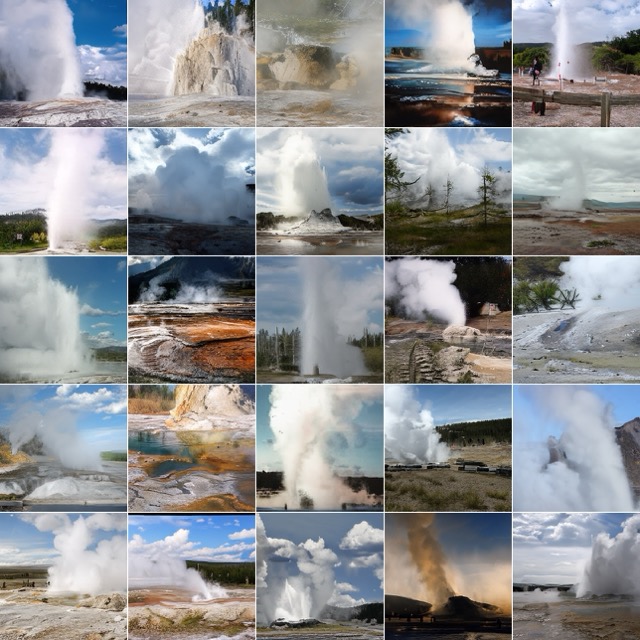}
    \end{minipage}
    \captionsetup{hypcap=false}
    \captionof{figure}{Uncurated class-conditional generation on ImageNet-256. Left: volcano (class 980). Right: geyser (class 974). Each grid contains 25 samples.}
    \label{fig:supp_gen_volcano_geyser}
\end{minipage}\par\medskip

\flushbottom
\section{\OURS Tokenizer Training Details}
\label{sec:tokenizer_training}

We train the \OURS tokenizer on the ImageNet-1K training set at a resolution of $256 \times 256$. The \textit{Encoder} adopts a Vision Transformer (ViT) backbone comprising 8x Transformer~\cite{transformer} blocks with a patch size of $16 \times 16$ and learnable positional embeddings~\cite{heo2024rotarypositionembeddingvision,su2023roformerenhancedtransformerrotary}. The \textit{Aggregator} and \textit{Decoder} modules follow a similar architecture, each containing 8x Transformer~\cite{transformer} blocks, but we only apply structural instruction embedding instead using standard position embedding widly using in Visiton Transformer. 

The hidden embedding dimension across all modules is fixed at $512$. Following LlamaGen~\cite{sun2024autoregressive}, we employ a shared codebook across all tree levels with vocabulary size $V = 16384$ and latent dimensionality $d = 8$. During training, probabilistic tree sampling is applied with child node expansion probability $\gamma = 0.5$, which balances computational efficiency and reconstruction fidelity. The tokenizer contains approximately $120$M parameters.

For the quadtree structure, we use two depth levels with spatial token grids of $8 \times 8$ and $16 \times 16$. The unpatched spatial footprint of each token is set to $p_1 = p_2 = 8$. The hidden embedding dimension is fixed to $512$ across all modules. Following LlamaGen~\cite{sun2024autoregressive}, we use a shared codebook across tree levels with vocabulary size $V = 16384$ and latent dimension $d = 8$. During training, probabilistic tree sampling is applied with child expansion probability $\gamma = 0.5$, balancing computational efficiency and reconstruction fidelity. In total, the tokenizer contains approximately $120$M parameters.

To optimize the tokenizer, we employ a combination of reconstruction and adversarial objectives. The overall training objective $\mathcal{L}_{total}$ is formulated as a weighted sum of four distinct loss components:$$\mathcal{L}_{total}=\lambda_{MSE}\mathcal{L}_{MSE}+\lambda_{LPIPS}\mathcal{L}_{LPIPS}+\lambda_{VQ}\mathcal{L}_{VQ}+\lambda_{GAN}\mathcal{L}_{GAN}$$where $\mathcal{L}_{MSE}$ denotes the Mean Squared Error for exact pixel-level fidelity, and $\mathcal{L}_{LPIPS}$ is the Learned Perceptual Image Patch Similarity loss, which preserves high-frequency visual details and structural integrity. $\mathcal{L}_{VQ}$ represents the Vector Quantization commitment loss necessary to align the continuous visual representations with the discrete codebook. To further enhance the perceptual realism and sharpness of the reconstructed images, we incorporate an adversarial GAN loss, denoted by $\mathcal{L}_{GAN}$. The hyperparameters $\lambda_{MSE}$, $\lambda_{LPIPS}$, $\lambda_{VQ}$, and $\lambda_{GAN}$ control the relative contribution of each corresponding objective. To ensure initial training stability and proper codebook convergence, the adversarial weight $\lambda_{GAN}$ is strictly set to 0 initially and is only activated after the first 200,000 training steps. For each loss weight, we apply the same setting with~\cite{miwa2025one}

\section{\OURS Generator Training Details}
\label{sec:generator_training}

The architecture of the \OURS generator follows the decoder-only Autoregressive(AR) generation paradigm adopted in LlamaGen~\cite{sun2024autoregressive}. We incorporate several modern architectural components, including Query-Key (QK) Normalization~\cite{henry2020querykeynormalizationtransformers}, SwiGLU activation functions~\cite{shazeer2020glu}, and Rotary Position Embeddings (RoPE)~\cite{heo2024rotarypositionembeddingvision}. We scale the two-level generator across four model sizes: 111M (B), 343M (L), 697M (XL), and 947M (XXL) parameters. The three-level generator contains 344M parameters. All models are trained for 300 epochs (approximately 360K iterations) using a global batch size of $1024$, without applying exponential moving average (EMA) to the model weights. Optimization is performed using the AdamW~\cite{kingma2014adam} optimizer with $\beta_1=0.9$, $\beta_2=0.95$, and weight decay $0.05$. The learning rate is set to $4\times10^{-4}$ for the first 250 epochs without warm-up, followed by a linear decay to $1\times10^{-5}$ over the final 50 epochs. To mitigate overfitting and enable Classifier-Free Guidance~\cite{ho2022classifierfreediffusionguidance,cfg} (CFG), the class conditioning embeddings are randomly dropped with probability $0.1$ during training.

During inference, we employ a standard Key-Value (KV) cache to accelerate autoregressive decoding, together with a dynamic power-cosine CFG scheduler to improve sample fidelity. A key characteristic of our framework is that the quadtree topology must be predefined before generation. 

\textit{Quadtree structure during inference.} For class-conditional ImageNet evaluation, we sample quadtree topologies independently of image content using a probabilistic expansion policy. Each sampled topology specifies the regions to refine, and its node instructions are presented in breadth-first order as the generator predicts visual codes.

\section{High-Resolution Reconstruction and Generation}
\label{sec:high_resolution_512}
We evaluate \OURS on ImageNet at $512\times512$ to test whether spatially adaptive tokenization and topology-conditioned generation extend to higher resolution. We retain two quadtree levels and enlarge the spatial grids to $16\times16$ and $32\times32$, separating increased image resolution from the deeper-tree experiment.

\paragraph{Reconstruction.}
Region-wise Complexity Guidance allocates tokens according to the reconstruction benefit of refining each region. \cref{tab:reconstruction_512} compares this allocation with fixed 2D grids and compact 1D tokenizers, demonstrating high-resolution reconstruction with fewer tokens than the 1,024-token grid baselines.

\begin{table}[t]
% \vspace{-2mm}
\centering
\small
\setlength{\tabcolsep}{4pt}
\caption{\textbf{ImageNet-1K reconstruction with $512\times512$ inputs.} \#Tokens is the mean for \OURS. $^{\dagger}$: LlamaGen resizes reconstructions to $256\times256$ for evaluation. MaskGIT and TiTok values follow Table~2 of the TiTok paper. -- denotes unreported values.}
\label{tab:reconstruction_512}
\begin{tabular*}{\textwidth}{@{\extracolsep{\fill}}llrrrr@{}}
\toprule
Type & Tokenizer & \#Tokens & Codebook size & rFID $\downarrow$ & PSNR $\uparrow$ \\
\midrule
\multirow{2}{*}{2D} & MaskGIT VQ-GAN~\cite{chang2022maskgit} & 1024 & 1024 & 1.97 & -- \\
 & LlamaGen$^{\dagger}$~\cite{sun2024autoregressive} & 1024 & 16384 & 0.70 & 23.03 \\
\midrule
\multirow{2}{*}{1D} & TiTok-L-64~\cite{yu2024imageworth32tokens} & 64 & 4096 & 1.77 & -- \\
 & TiTok-B-128~\cite{yu2024imageworth32tokens} & 128 & 4096 & 1.52 & -- \\
\midrule
Quadtree & \textbf{\OURS} & 987 & 16384 & 0.84 & 24.96 \\
\bottomrule
\end{tabular*}
% \vspace{-3mm}
\end{table}

\paragraph{Generation.}
The 344M generator synthesizes native-$512\times512$ images under random topologies supplied before generation. We evaluate 50,000 class-conditional samples using the ImageNet-512 ADM reference. In \cref{tab:generation_512}, \OURS achieves lower reported gFID than DiT-XL/2 and CAT with a smaller generator, supporting the extension to high-resolution synthesis. These are system-level comparisons with different tokenizers and training and sampling settings.

\begin{table}[t]
\vspace{-2mm}
\centering
\small
\setlength{\tabcolsep}{4pt}
\caption{\textbf{Class-conditional ImageNet generation at $512\times512$.} Size denotes generator parameters. CAT uses a continuous VAE with DiT ($c=8$, original Table~7); TiTok uses MaskGIT with 64 sampling steps. VAR uses next-scale prediction, while \OURS uses causal autoregressive decoding. Results are system-level comparisons; -- denotes unreported metrics.}
\label{tab:generation_512}
\begin{tabular*}{\textwidth}{@{\extracolsep{\fill}}lrrrrr@{}}
\toprule
Model & Size & gFID $\downarrow$ & IS $\uparrow$ & Precision $\uparrow$ & Recall $\uparrow$ \\
\midrule
DiT-XL/2~\cite{peebles2023scalable} & 675M & 3.04 & 240.82 & 0.84 & 0.54 \\
CAT~\cite{shen2025cat} & $\approx$431M & 4.38 & 181.03 & 0.76 & 0.48 \\
TiTok-B-128~\cite{yu2024imageworth32tokens}-MaskGIT & 177M & 2.13 & 261.2 & -- & -- \\
VAR-$d36$-s~\cite{tian2024visual} & 2.3B & 2.63 & 303.2 & -- & -- \\
\midrule
\textbf{\OURS} & 344M & 2.734 & 268.01 & 0.835 & 0.537 \\
\bottomrule
\end{tabular*}
\vspace{-3mm}
\end{table}

\subsection{Experimental details}
\label{sec:high_resolution_512_details}
The 512 model uses two token levels on $16\times16$ and $32\times32$ grids. All 256 coarse tokens are retained; expanding a coarse node adds four fine tokens, giving a maximum of 1,280 tokens.

\paragraph{Tokenizer configuration and reconstruction.}
The tokenizer configuration uses a ViT-small encoder, a ViT-large decoder, a patch size of 16, and a shared 16,384-entry codebook with 8-dimensional codes. The training recipe uses a coarse-node expansion probability of 0.73, $512\times512$ random crops, and random horizontal flips. AdamW uses learning rate $10^{-4}$, $(\beta_1,\beta_2)=(0.9,0.999)$, and weight decay $10^{-4}$. The cosine schedule has 10,000 warm-up steps and final learning rate $10^{-5}$. The configured training budget is 350,000 steps with global batch size 512, BF16 precision, EMA, and gradient clipping at 1.0. The reconstruction, perceptual, VQ, and adversarial loss weights are 1.0, 1.1, 1.0, and 0.1, respectively; adversarial training starts at step 200,000.

\paragraph{Generator configuration and evaluation.}
We evaluate the 344M generator checkpoint at 400,000 steps, conditioned on a topology supplied before generation with a $16\times16$ coarse grid and expansion probability 0.75. Sampling uses linear classifier-free guidance with a final scale of 16, temperature 1.0, and BF16 precision. We generate 50,000 native-$512\times512$ images, with 50 samples per ImageNet class, and evaluate them against the ImageNet-512 ADM reference. The mean generation sequence length is 1,021.28 tokens.

\clearpage
\section{Efficiency Measurement Details}
\label{sec:efficiency_details}

\paragraph{Generation efficiency.}
\cref{tab:generation_efficiency} compares \OURS with LlamaGen at matched model scales on an A100-80GB GPU, using BF16, KV caching, and matched batch sizes and CFG modes. At the XL scale, LlamaGen's training hyperparameters were unstable in our setting, motivating the 697M backbone used for quality evaluation. Timing instead uses a 775M \OURS model to match LlamaGen-XL. Lower token budgets yield larger speedups, while the near-equal-token gains reflect implementation differences. The comparison measures runtime at specified budgets; \cref{fig:expansion_fid} examines the associated quality tradeoff.

\begin{table}[t]
\vspace{-2mm}
\centering
\caption{\textbf{Generation speedup relative to LlamaGen.} The LlamaGen timing baseline uses randomly initialized generator weights in the $256\times256$ configuration and generates 256 tokens. At $p=0.25$ and $0.50$, ranges span batch sizes 1/32 and CFG scales 1/2. The $p=0.75$ column reports the near-equal-token comparison. Ratios measure autoregressive generation, excluding image decoding; values above 1 indicate faster generation. The 775M comparison is timing-only (\cref{sec:efficiency_details}).}
\label{tab:generation_efficiency}
\small
\setlength{\tabcolsep}{10pt}
\begin{tabular}{lccc}
\toprule
Parameters & $p=0.25$ & $p=0.50$ & $p=0.75$ \\
Approx. tokens & 128 & 192 & 255 \\
\midrule
111M & 1.81--2.40$\times$ & 1.14--1.58$\times$ & 1.20$\times$ \\
343M & 1.62--1.94$\times$ & 1.09--1.20$\times$ & 1.09$\times$ \\
775M & 1.57--1.91$\times$ & 1.15--1.21$\times$ & 1.10$\times$ \\
\bottomrule
\end{tabular}
\vspace{-3mm}
\end{table}

\paragraph{Generation protocol.}
Each matched-size pair runs on one NVIDIA A100-SXM4-80GB GPU with PyTorch 2.7.1 and CUDA 12.8. Both implementations use BF16, a KV cache, full-vocabulary multinomial sampling, and the same batch size and CFG execution mode. We measure 50 batches after 10 warmups. Generation latency covers autoregressive token sampling; generation-plus-decoding latency additionally includes the tokenizer decoder. Checkpoint loading and disk/network I/O are excluded. Reported times are per batch, not per image. The two-level random topology retains 64 coarse nodes and adds four children for each independently expanded coarse node, giving an expected $64+256p$ tokens per image.

The benchmark uses 111M, 343M, and 775M generators. For latency measurements, we instantiate the corresponding LlamaGen architectures with randomly initialized generator weights and the $256\times256$ configuration, using 256 autoregressive decoding steps. This benchmark measures runtime under the specified architecture and sampling settings; image quality is evaluated separately in \cref{tab:main_results}. \cref{tab:generation_efficiency} summarizes generation speedups at $p=0.25$, $0.50$, and $0.75$, while \cref{tab:generation_latency_details} reports absolute latencies for all three settings at batch size 32 and CFG scale 2.

\begin{table}[htbp]
\vspace{-2mm}
\centering
\caption{\textbf{Absolute generation latency at batch size 32 and CFG scale 2.} Means over 50 timed batches, in seconds per batch. Gen. denotes autoregressive sampling; + decode includes image decoding. Results cover $p=0.25$, $0.50$, and $0.75$ under the protocol in Appendix~\ref{sec:efficiency_details}. Speedup is the ratio of LlamaGen to \OURS generation latency.}
\label{tab:generation_latency_details}
\small
\setlength{\tabcolsep}{6pt}
\begin{tabular}{llrrrr}
\toprule
Scale & Method & Tokens & Gen. (s) & + decode (s) & Speedup \\
\midrule
111M & LlamaGen & 256.0 & 5.440 & 5.551 & 1.00$\times$ \\
111M & \OURS ($p=0.25$) & 126.6 & 2.524 & 2.560 & 2.16$\times$ \\
111M & \OURS ($p=0.50$) & 189.2 & 3.528 & 3.570 & 1.54$\times$ \\
111M & \OURS ($p=0.75$) & 255.0 & 4.533 & 4.586 & 1.20$\times$ \\
\midrule
343M & LlamaGen & 256.0 & 8.404 & 8.512 & 1.00$\times$ \\
343M & \OURS ($p=0.25$) & 126.6 & 4.656 & 4.692 & 1.80$\times$ \\
343M & \OURS ($p=0.50$) & 189.2 & 7.504 & 7.550 & 1.12$\times$ \\
343M & \OURS ($p=0.75$) & 255.0 & 7.710 & 7.763 & 1.09$\times$ \\
\midrule
775M & LlamaGen & 256.0 & 12.612 & 12.720 & 1.00$\times$ \\
775M & \OURS ($p=0.25$) & 126.6 & 6.587 & 6.620 & 1.91$\times$ \\
775M & \OURS ($p=0.50$) & 189.2 & 10.643 & 10.688 & 1.18$\times$ \\
775M & \OURS ($p=0.75$) & 255.0 & 11.465 & 11.520 & 1.10$\times$ \\
\bottomrule
\end{tabular}
\vspace{-3mm}
\end{table}

\paragraph{Pretokenization latency.}
Adaptive allocation adds image-dependent search to pretokenization. \cref{tab:pretokenization_efficiency} measures latency from GPU image tensors to discrete codes, including tree search and LPIPS-based selection but excluding data loading, transfers, and reconstruction decoding. This pipeline is slower than fixed-token encoding, but its cost is incurred only during data preparation and can be amortized by caching codes for generator training; generation under a supplied topology requires no image-dependent search.

\begin{table}[t]
\vspace{-2mm}
\centering
\caption{\textbf{Pretokenization latency at $256\times256$.} Mean milliseconds per image on one A100-80GB GPU. Token counts for \OURS are measured on the timed batches (32 / 128); fixed encoders use a constant budget. Timing includes adaptive selection for \OURS and ends at discrete codes, before reconstruction decoding.}
\label{tab:pretokenization_efficiency}
\small
\setlength{\tabcolsep}{9pt}
\begin{tabular}{lrrr}
\toprule
Tokenizer & Tokens/image & Batch 32 & Batch 128 \\
\midrule
LlamaGen VQ-16 & 256 & 2.927 & 2.885 \\
TiTok L-32 & 32 & 1.480 & 1.352 \\
TiTok B-64 & 64 & 0.586 & 0.512 \\
TiTok S-128 & 128 & 0.305 & 0.277 \\
\midrule
\OURS & 234.3 / 233.2 & 8.795 & 6.186 \\
\bottomrule
\end{tabular}
\vspace{-3mm}
\end{table}

\paragraph{Pretokenization protocol.}
We use a fixed ImageNet validation prefix and time five warmups followed by 30 repeats at batch sizes 32 and 128. Inputs are preloaded $256\times256$ image tensors on the GPU; outputs are discrete token indices on the GPU. Dataset decoding, host-to-device transfer, disk I/O, and final image reconstruction decoding are excluded. The optimized \OURS path includes adaptive-tree search, LPIPS-based selection, and code extraction. LlamaGen VQ-16 and the released TiTok L-32, B-64, and S-128 variants use fixed-token encoders. The comparison measures actual preprocessing pipelines, not matched encoder architectures. At batch size 128, \OURS processes 161.66 images/s versus 346.58 for LlamaGen VQ-16; the additional cost supports image-dependent regional allocation. The measured mean token counts differ slightly from dataset-wide reconstruction averages because the timing uses a fixed validation prefix.

\section{Spatial Layout Evaluation Details}
\label{sec:layout_eval_details}

We evaluate the frozen two-level ImageNet-trained generator and tokenizer using CFG scale 6, a power-cosine guidance schedule with exponent 1, and sampling temperature 1. Each condition contains 20 images for each of 50 fixed ImageNet classes, giving 1,000 images per condition.

The four prescribed layouts correspond to the left, right, top, and bottom half-planes. Each topology retains all nodes of an $8\times8$ coarse grid and expands the 32 cells in the target half-plane, adding four children per cell for a total of 192 visual tokens. The matched random baseline uniformly selects exactly 32 coarse cells for expansion independently for each image. Its generated image set is shared across the four directional evaluations.

Grad-CAM uses the intended class in ResNet-50. Box localization uses the highest-confidence intended-object box from Grounding DINO, and SAM~2 receives that box as its mask prompt. The Grad-CAM center of mass, box center, or mask centroid is tested against the requested half-plane. Box and mask scores count missed detections as failures and use all generated images as the denominator.

FID-50cls compares Inception-v3 features of the 1,000 generated images in each condition with the same 2,500 ImageNet validation images from the 50 evaluated classes. Its value is 35.03 for the random-topology baseline and 35.15, 34.88, 34.97, and 35.05 for the prescribed left, right, top, and bottom layouts, respectively.

\section{Complexity Guidance: Metric and Threshold}
\label{sec:guidance_ablation}
We ablate the probing signal and its expansion threshold $\tau$ (\cref{tables/ablation_complex}), comparing MSE, LPIPS~\cite{zhang2018unreasonableeffectivenessdeepfeatures}, and their linear combination ($\mathcal{L}_\mathrm{MSE}\times10+\mathcal{L}_\mathrm{LPIPS}$). At a matched $\sim$230-token budget, LPIPS aligns allocation with perceptual quality better than MSE or the combination (rFID $1.46$ vs.\ $1.67$ and $1.58$); sweeping $\tau$ shows $0.05$ best balances token count and fidelity. We therefore adopt LPIPS with $\tau{=}0.05$ as the default.

\begin{table}[htbp]
\vspace{-2mm}
    \caption{\textbf{Ablation of complexity guidance metric and threshold $\tau$.} We compare MSE, LPIPS, and their linear combination as probing metrics; for LPIPS we sweep the expansion threshold $\tau$, and report MSE and the combination at a matched $\sim$230-token budget on ImageNet-1K. LPIPS at $\tau{=}0.05$ achieves the best perceptual quality (rFID $1.46$) at $230$ tokens, and is adopted as the default.}
    \label{tables/ablation_complex}
    \centering
    \setlength{\tabcolsep}{2em}
    \resizebox{0.9\linewidth}{!}{%
        \begin{tabular}{@{\hspace{1em}} l c c c c c @{\hspace{0em}}}
            \toprule
            \textbf{Guidance Metric} & \textbf{Threshold ($\tau$)} & \textbf{\#Tokens} $\downarrow$ & \textbf{rFID} $\downarrow$ & \textbf{PSNR} $\uparrow$ \\
            \midrule
            $\mathcal{L}_\mathrm{MSE}$ & 0.0005 & 230 & 1.67 & 20.52 \\
            \midrule
            \multirow{3}{*}{$\mathcal{L}_\mathrm{LPIPS}$}
                                     & 0.03                        & 267                            & 1.46                       & 20.24                    \\
                                     & \textbf{0.05}               & \textbf{230}                   & \textbf{1.46}              & \textbf{20.37}           \\ % Highlighted your best setting
                                     & 0.06                        & 195                            & 1.75                       & 19.99                    \\
            \midrule
            Combined ($\mathcal{L}_\mathrm{MSE}\times10+\mathcal{L}_\mathrm{LPIPS}$) & 0.06 & 230 & 1.58 & 20.37 \\
            \bottomrule
        \end{tabular}
    }
\vspace{-3mm}
\end{table}

\section{Limitations and Future Work}
\label{sec:limitations}

While \OURS demonstrates strong potential in dynamic token allocation and structured autoregressive generation, our current study has two primary limitations. 

\textbf{Scale and Modality Constraints.} First, due to computational resource limitations, our empirical evaluations are currently confined to class-conditional image generation on the ImageNet dataset. We have not yet scaled the framework to massive, open-domain datasets (e.g., LAION) nor extended the architecture to encompass complex text-to-image synthesis paradigms. Scaling the model capacity and extending the cross-attention mechanisms for open-vocabulary text conditioning require significantly more training compute and represent a critical next step.

\textbf{Dataset Bias in Layout Control.} Second, the inherent bias of the ImageNet dataset towards object-centric, single-subject compositions restricts the upper bound of our zero-shot spatial layout control. Although \OURS effectively localizes primary subjects and separates them from the background based on topological priors, synthesizing highly complex, multi-object scenes with intricate relational layouts remains challenging. The model's structural prior is naturally limited by the compositional diversity of its training data.

We leave the scaling of \OURS to large-scale text-to-image generation and its application to scene-level dataset for complex, multi-object layout control to future work.

\end{document}